\documentclass{article}

\usepackage{caption}
\usepackage{subfigure}
\usepackage{url}
\usepackage{multirow}
\usepackage{bm}
\usepackage{spconf,amsmath,graphicx}
\usepackage{times}
\usepackage{latexsym}
\usepackage[T1]{fontenc}
\usepackage[utf8]{inputenc}
\usepackage{microtype}
\usepackage{inconsolata}
\usepackage{multirow}
\usepackage{graphicx}
\usepackage{booktabs}
\usepackage{CJKutf8}
\usepackage{array}
\usepackage{float}
\usepackage{url}
\usepackage{xcolor}
\usepackage{hyperref}

\usepackage{xpinyin}
\usepackage{enumerate}

\title{Joint Training And Decoding for Multilingual End-to-End Simultaneous Speech Translation}
\def\star{$^*$}

\name{Author(s) Name(s)}
\address{Author Affiliation(s)}

\name{Wuwei Huang$^{1}$\star\thanks{$^*$ The work was done while the first author was a postgraduate student at Tianjin University.}\ \ \ \ Renren Jin$^{2}$\ \ \ \ Wen Zhang$^{1}$\ \ \ \ Jian Luan$^{1}$\ \ \ \  Bin Wang$^{1}$\ \ \ \ Deyi Xiong$^{2}$$\dagger$\thanks{$\dagger$ Corresponding author.}} 
\address{$^{1}$Xiaomi AI Lab, Beijing, China \qquad \\$^{2}$College of Intelligence and Computing, Tianjin University, Tianjin, China}
\begin{document}
\maketitle
\begin{abstract}
Recent studies on end-to-end speech translation(ST) have facilitated the exploration of multilingual end-to-end ST and  end-to-end simultaneous ST. In this paper, we investigate  end-to-end simultaneous speech translation in a one-to-many multilingual setting  which is closer to applications in real scenarios. We explore a separate decoder architecture and a unified  architecture for joint synchronous training in this scenario. To further explore knowledge transfer across languages, we propose an asynchronous training strategy on the proposed unified decoder architecture. A  multi-way aligned multilingual end-to-end ST dataset was curated as a benchmark testbed to evaluate our methods. Experimental results
demonstrate the effectiveness of our models on the collected dataset. Our codes and data are available at: \url{https://github.com/XiaoMi/TED-MMST}.
\end{abstract}

\begin{keywords}
End-to-End ST, Simultaneous Machine Translation, Multilingual\end{keywords}

\section{Introduction}
End-to-end speech translation (ST) directly translates speech utterances in  source language into texts in target language. Due to its advantages over cascade ST that suffers from error propagation and information loss ~\cite{DBLP:conf/acl/SperberP20}, end-to-end ST has recently attracted growing attention and made substantial progress ~\cite{DBLP:conf/acl/HuangWX21}.

Such progress has further promoted the exploration of end-to-end ST in many realistic scenarios. One of them is multilingual scenario, where speech utterances in one or multiple source languages are translated into multiple target languages, e.g., in online lectures or meetings. Inspired by multilingual neural machine translation (NMT), where a single learning system is trained for multiple language pairs \cite{DBLP:conf/naacl/FiratCB16,DBLP:conf/icassp/VuIPMSB14}, multilingual end-to-end ST has been investigated \cite{DBLP:conf/asru/InagumaDKW19,DBLP:conf/asru/GangiNT19}. The nature of multi-task learning and transfer learning across languages substantially benefit multilingual 
 end-to-end ST, especially for low-resource languages.

Yet another real-world scenario for end-to-end ST is simultaneous translation. Simultaneous speech translation begins to generate target words before the entire speech input is received \cite{DBLP:conf/ijcnlp/MaPK20,DBLP:conf/icassp/MaWDKP21}, requiring a trade-off between translation quality and latency. 
In this paper, we investigate both multilinguality and simultaneousness in end-to-end ST, i.e., one-to-many multilingual end-to-end simultaneous ST, which finds its applications in various scenarios, e.g., international conferences, online multilingual conversation, online lectures for students from different countries. These scenarios typically require simultaneously translating speech signals in one source language into multiple target languages. The combination of multilinguality and simultaneousness in end-to-end ST is nontrivial as it is confronted with challenges from both data scarcity and simultaneously decoding among different languages. We investigate two neural architectures for joint multilingual end-to-end simultaneous ST, a separate decoder model and a unified encoder-decoder model in a joint synchronous training. Specifically, the separate decoder model uses different decoders for different target languages, while the unified model shares all decoders across all target languages. To further explore knowledge transfer, we devise joint asynchronous training. 

Our contributions can be summarized as follows:
\begin{itemize}
\item[$\bullet$]We  present two models in a  synchronous training for multilingual end-to-end simultaneous ST. 
\end{itemize}
\begin{itemize}
\item[$\bullet$]In addition to standard  synchronous training, we further propose asynchronous training to explore knowledge transfer between different languages.
\end{itemize}
\begin{itemize}
\item[$\bullet$]We curate a dataset as a benchmark for multilingual end-to-end simultaneous ST.
\end{itemize}
\begin{itemize}
\item[$\bullet$]Our experiments demonstrate the effectiveness of the proposed framework and training strategy for multilingual end-to-end simultaneous speech translation.
\end{itemize}
\begin{figure*}[htbp]
\centering
\subfigure[Separate Decoder Model]
{
    \begin{minipage}{0.50\linewidth}
    \centering
    \includegraphics[height=6cm,width=8.5cm]{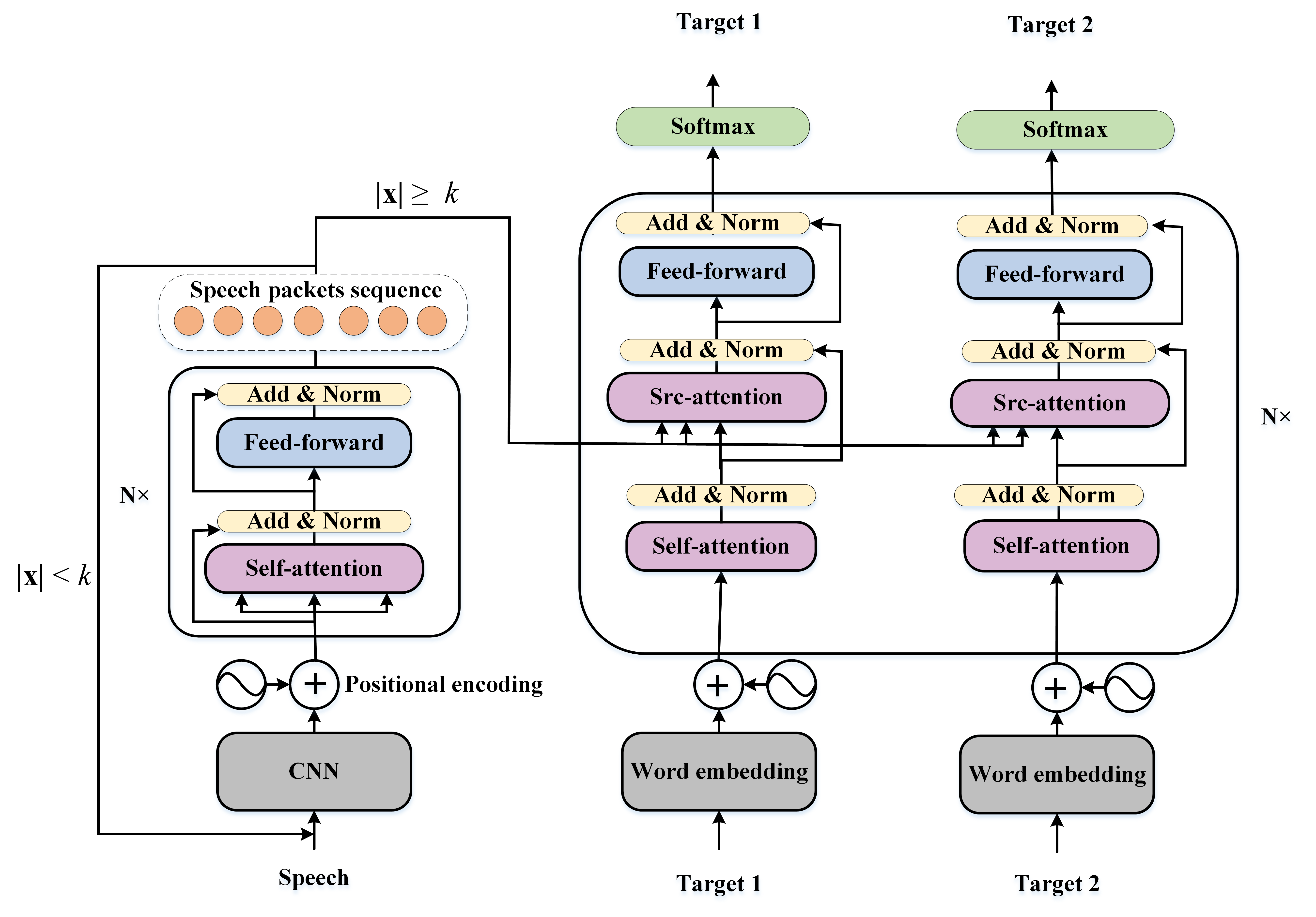}
    \end{minipage}
}
\subfigure[Unified Model]
{
    \begin{minipage}{0.47\linewidth}
    \centering
    \includegraphics[height=6cm,width=7cm]{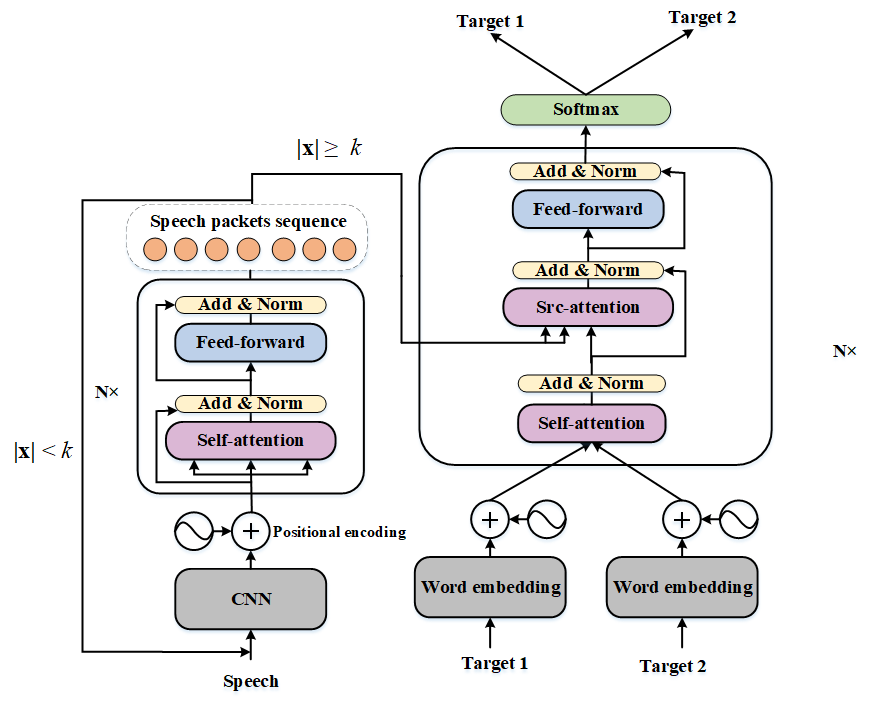}
    \end{minipage}
}
\caption{Diagram of joint multilingual end-to-end simultaneous speech translation.}
\label{fig1}
\end{figure*}
\vspace{-0.3cm}
\section{Related Work}
Since the pioneering works on end-to-end ST \cite{DBLP:journals/corr/BerardPSB16,DBLP:journals/corr/abs-2009-09704}, a wide range of methods have been proposed. \cite{DBLP:conf/interspeech/WeissCJWC17} and  \cite{DBLP:conf/icassp/BerardBKP18} take advantages of multitask learning to train end-to-end speech translation model by using automatic speech recognition (ASR) or machine translation (MT) as auxiliary tasks. To exploit rich resources in ASR and MT,  \cite{DBLP:conf/icassp/JiaJMWCCALW19} synthesizes training data for end-to-end speech translation from ASR and MT datasets.  \cite{DBLP:conf/acl/SaleskySB19}and \cite{DBLP:conf/emnlp/ZhangTHS20} optimize speech input modeling in the encoder by reducing the length of acoustic sequences. Knowledge distillation\cite{DBLP:conf/interspeech/LiuXZHWWZ19} and curriculum learning \cite{DBLP:conf/acl/WangWLZY20} have also been explored in end-to-end ST.

To overcome the issue of data scarcity, \cite{DBLP:conf/asru/GangiNT19} mixes data of different languages pairs to train a one-to-many multilingual end-to-end ST model.  \cite{DBLP:conf/asru/InagumaDKW19} investigates both one-to-many and many-to-many multilingual end-to-end ST.

Studies in simultaneous machine translation greatly inspire the exploration of simultaneous speech translation. \cite{DBLP:conf/ijcnlp/MaPK20} adapts methods originally proposed for simultaneous machine translation to end-to-end simultaneous ST. 
None of the aforementioned previous studies on either multilingual end-to-end ST or end-to-end simultaneous ST investigate the nontrivial conjunction of these two technologies.
The most recent work similar to ours is \cite{DBLP:conf/interspeech/XueW0PG22}, but it trains models neither synchronously nor asynchronously.
 \vspace{-0.3cm}
\section{Method}
We propose two joint training architectures for multilingual end-to-end simultaneous ST, which vary the degree to which parameters are shared across languages, as shown in Figure \ref{fig1}. Without loss of generality, we take two target languages  as a special case to discuss our methods.
\subsection{Separate Decoder Model}
Our separate decoder model for  multilingual end-to-end simultaneous ST shares speech encoder parameters across different languages. The training loss for the  $j^{th}$ target language is:
\begin{equation}\label{equ1}
\small
\mathcal{L}^{j}\left(\bm{\mathrm{x}},\bm{\mathrm{y}}^{j};\bm{\theta}_{e},\bm{\theta}_d^{j}\right) =  {\sum^{|\bm{\mathrm{y}}^{j}|}_{t=1}}\textnormal{-}\log p\left(\bm{\mathrm{y}}_t^{j}|\bm{\mathrm{y}}_{<t}^{j},\bm{\mathrm{x}}_{<g(t)};\bm{\theta}_{e},\bm{\theta}_d^{j}\right)
\end{equation}
where $\bm{\theta}_{e}$ and $\bm{\theta}_d^{j}$ represent the parameters of the encoder and the $j^{th}$ decoder, respectively.
$\bm{\mathrm{y}}^{j}$ means the ground-truth translation corresponding to the $j^{th}$ target language, and $\bm{\mathrm{y}}_{<t}$ is the first $t\textnormal{-}1$ tokens of the $j^{th}$ target translation. Decoders for different languages do not share parameters. $g(t)$ denotes the length of source speech packets read by the model before writing the $t^{th}$ target token. The separate decoder  model is similar to \cite{DBLP:conf/coling/LePWGSB20}, which explores interactive decoding between two tasks or two target languages. Our motivation is significantly different from theirs. We aim at cross-lingual knowledge transfer, rather than using the information of another task or another language for decoding.
\subsection{Unified Model}
The second model for multilingual end-to-end simultaneous ST shares all parameters across different languages in a unified model. To enable the model to distinguish different target languages during decoding,  a token identifying the target language is prepended to the target sequence of each training instance. The training loss for the unified model for different languages can be formulated as follows
\begin{equation}\label{equ2}
\small
\mathcal{L}\left(\bm{\mathrm{x}},\bm{\mathrm{y}};\bm{\theta}_{e},\bm{\theta}_d\right) =  {\sum^{|\bm{\mathrm{y}}|}_{t=1}}\textnormal{-}\log p\left(\bm{\mathrm{y}}_t|\bm{\mathrm{y}}_{<t},\bm{\mathrm{x}}_{<g(t)};\bm{\theta}_{e},\bm{\theta}_d\right)
\end{equation}
where the decoders for different languages share the same parameters $\bm{\theta}_{d}$.

\subsection{Joint Training}
The standard training \emph{(i.e., joint synchronous training)} is used for both the separate decoder model and the unified model. We also propose a novel training strategy: \emph{joint asynchronous training}.
Without loss of generality, we take two target languages as an example.
\vspace{-0.3cm}
\subsubsection{Joint Synchronous Training} \label{subsubsec:jst}
To deal with simultaneousness, we adopt wait-$k$ strategy and  fixed pre-decision module~\cite{DBLP:conf/ijcnlp/MaPK20}.
Our model first reads $k$ speech packets, each of which contains fixed $q$ speech frames where $q$ is a hyper-parameter in the fixed pre-decision module. The decoder begin to generates one token simultaneously for each language after $k$ speech packet is read.
Since the same $k$ value is employed on different target languages, we call this training method \emph{joint synchronous training}.
Once a speech packet is accepted, the operation taken by the decoder is formalized as follows:
\begin{equation}\label{equ3}
 \bm{\mathrm{Operation}}=\left\{
 \begin{array}{lcr}
 \mathrm {continue\ to\ read} & {|\bm{\mathrm{x}}|<k}\\
 \mathrm{output}\ \bm{\mathrm{y}}_{t}^{1},\ \bm{\mathrm{y}}_{t}^{2} & {|\bm{\mathrm{x}}|\geq k}\\
 \end{array}
 \right.
\end{equation}
where $\bm{y}_{t}^{1}$ and $\bm{y}_{t}^{2}$ represent the $t^{th}$ token of each target language, and $|\bm{\mathrm{x}}|$ represent the length of speech packet.
Our final training objective is:
\begin{equation}
\begin{aligned}
\label{equ4}
 \mathcal{L}(\bm{\mathrm{x}},\ &\bm{\mathrm{y}}^{1},\bm{\mathrm{y}}^{2};\bm{\theta}_{e},\bm{\theta}_d^{1},\bm{\theta}_d^{2})=\\ &\mathcal{L}^{1}\left(\bm{\mathrm{x}},\bm{\mathrm{y}}^{1};\bm{\theta}_{e},\bm{\theta}_d^{1}\right) + \mathcal{L}^{2}\left(\bm{\mathrm{x}},\bm{\mathrm{y}}^{2};\bm{\theta}_{e},\bm{\theta}_d^{2}\right)
 \end{aligned}
\end{equation}
$\mathcal{L}^{1}$ and $\mathcal{L}^{2}$ are the loss functions for each language, as defined by Eq.~\ref{equ1}. 
 \vspace{-0.3cm}
\subsubsection{Joint Asynchronous Training}
As introduced in the previous subsection, multiple languages use the same $k$ during synchronous training. In actual usage scenarios, each target language holds an appropriate $k$ to balance the translation quality and latency. In many cases, these $k$ values are not identical across languages. We hence propose an asynchronous training strategy.

Once a speech packet is received, the subsequent operation of the decoder can be defined as follows:
\begin{equation}\label{equ5}
 \bm{\mathrm{Operation}}=\left\{
 \begin{array}{lcr}
  \mathrm{continue \ to \ read}\ &{|\bm{\mathrm{x}}|<k^{1}, k^{2}}\\
 \mathrm{output}\ \bm{\mathrm{y}}_{t}^{1}\ &{k^{1}\leq |\bm{\mathrm{x}}| < k^{2}}\\
 \mathrm{output}\ \bm{\mathrm{y}}_{t}^{1},\ \bm{\mathrm{y}}_{t}^{2}\ &{|\bm{\mathrm{x}}|\geq k^{1},k^{2}}\\
 \end{array}
 \right.
\end{equation}
where $k^{1}$ and $k^{2}$ are  the $k$ values corresponding to the two languages, $k^{1}$ is smaller than $k^{2}$.

In order clearly describe  the process of asynchronous translation, 
we compare asynchronous training with synchronous training.
In the asynchronous schema, we assume the case of $k$=4 for English-to-Spanish and $k$=6 for English-to-French, while in synchronous schema, we set $k$ to 6. The decoder in the asynchronous schema starts to generate  a Spanish translation with a delay of  4 speech packets, and starts to generate a French translation with a delay of 6 speech packets. In contrast to the decoder in the asynchronous schema, the decoder in the synchronous schema begins to generate  Spanish and French translations simultaneously with a delay of 6 speech packets.

\section{Data Curation}
Most previous simultaneous speech translation research is based on the MuST-C \cite{di-gangi-etal-2019-must} dataset, which comes from TED talks \cite{long-etal-2020-ted,long2020shallowdiscourseannotationchinese}. Unfortunately MuST-C does not include data where  source speech signal is translated into multiple target languages. Thus, we curate a multilingual simultaneous speech translation dataset from TED website.

TED website offers English talks on various topics, as well as translations into different languages. In this paper, we use French and Spanish as  target languages for simultaneous training. To achieve this, we crawled approximately 2.5K  speech audios   and the corresponding English transcripts, French and Spanish translated texts from TED. After collecting the raw data, a two-step alignment should be performed to meet the training requirements. First, we preformed source sentence segmentation based on hard punctuation, using Gentle\footnote{\url{https://github.com/lowerquality/gentle}} to achieve audio-to-transcript alignments. Second, we created alignments that are unique to each En-XX language pair by Laser\footnote{\url{https://github.com/facebookresearch/LASER}} and finally obtained the intersection of the alignmnets across target texts. We split the final $175$K sentences into $171$K sentences as the training set, $1.5$K as the dev set and $2.7$K as the test set. The details of the dataset are shown in Table \ref{table1}. We also conducted offline experiments and  the BLEU score for En-Es and En-Fr are 28.56 and 24.46, respectively.
 \vspace{-0.3cm}
\section{Experiments}
We conducted experiments to evaluate the effectiveness of the proposed models on the curated dataset.
\begin{table}[t]
\renewcommand\arraystretch{1.3}
\centering
\scalebox{0.9}{\begin{tabular}{c|c|c|c|c|c} %
\hline
Domain & Hours & \multicolumn{3}{|c|}{Sentences} & Direction \\
\hline
\multirow{2}*{TED}&\multirow{2}*{$350$} & train & dev & test &\multirow{2}*{En$\rightarrow${Fr, Es}} \\
& & $171$K & $1.5$K & $2.7$K & \\
\hline
\end{tabular}}
\caption{Statistics of the curated data for multilingual end-to-end simultaneous speech translation.}\label{table1}
\end{table}
\vspace{-0.3cm}
\subsection{Evaluation Metrics}
Typically, simultaneous speech translation is evaluated with both translation quality and latency. The former was measured with detokenized BLEU \cite{DBLP:conf/acl/PapineniRWZ02}  in our experiments. In terms of latency, average latency (AL) \cite{DBLP:conf/acl/MaHXZLZZHLLWW19}, average proportion (AP) \cite{DBLP:journals/corr/ChoE16} and  Differentiable Average Lagging (DAL) \cite{DBLP:conf/ijcnlp/MaPK20} are commonly used. Under the simultaneous mechanism adopted by this paper, the latency is positively correlated with $k$. In order to represent the results of asynchronous experiments more clearly, we used $k$ as evaluation metric for latency throughout this paper.

\vspace{-0.3cm}
\subsection{Settings}
We built our models based on the Fairseq\footnote{\url{https://github.com/pytorch/fairseq}} toolkit. 
We extracted 80-dimensional Fbank features from audio files.
We used a shared vocabulary  with a size of $8K$ tokens and fixed the vocabulary for all experiments.
The dimension of the attention was set to $256$. We used $12$-layer encoder and the speech encoder are initialized
based on a pre-trained ASR task. The hyper-parameter $q$ introduced in Section~\ref{subsubsec:jst} wet set to $7$. The number of decoder layers  was set to $6$. We trained $70$ epochs for each model and used the Adam optimizer. All models were run on four NVIDIA RTX A6000 GPUs. After training, we test our model on SimulEval\footnote{\url{https://github.com/facebookresearch/SimulEval}} with greedy decoding strategy.

\begin{table}[t]
\renewcommand\arraystretch{1.215}
\centering
\scalebox{0.9}{\begin{tabular}{c|c|c|c|c|c} %
\hline
Tasks & Models & $k$=$3$ & $k$=$4$ & $k$=$5$ & $k$=$6$\\
\hline
\multirow{3}*{En$\rightarrow$Es} & Bilingual & $7.90$ & $11.13$ & $12.82$ & $15.80$ \\
& Separate Decoder & $9.47$ & $12.77$ & $15.62$ & $17.39$ \\
& Unified & $11.09$ & $12.62$ & $15.89$ & $17.55$ \\
\hline
\multirow{3}*{En$\rightarrow$Fr} & Bilingual & $10.09$ & $13.71$ & $15.59$ & $16.87$ \\
& Separate Decoder & $11.57$ & $14.20$ & $16.66$ & $17.67$ \\
& Unified & $13.39$ & $14.69$ & $17.35$ & $18.19$ \\
\hline
\end{tabular}}
\caption{BLEU scores on En-Es and En-Fr corresponding to $k$ in SimulEval with 3, 4, 5, 6, respectively.}\label{table2}
\end{table}





\begin{table}[t]
\renewcommand\arraystretch{1.0}
\centering
\begin{tabular}{c|c|c|c|c} %
\hline
 Training Strategy & $k_{Es}$ & $k_{Fr}$ & En$\rightarrow$Es & En$\rightarrow$Fr \\
\hline

\multirow{2}*{Synchronous} & 4 & 4 & 12.62 & 14.69\\
& 6 & 6 & 17.55 & 18.19 \\

\hline
\multirow{2}*{Asynchronous} & 6 & 4 & 17.77 & 15.74\\
& 4 & 6 & 14.40 & 18.61 \\
\hline
\end{tabular}
\caption{BLEU scores of asynchronous translation on En-Es and En-Fr.}\label{table3}
\end{table}
 
\section{Results and Analyses}
\subsection{Performance of the Separate Decoder Model and Unified Model}
Table \ref{table2} shows the comparison results of the proposed separate decoder model and unified model against the standard bilingual translation model with varying $k$. It is clear that both proposed models for multilingual end-to-end simultaneous ST are able to  significantly improve  translation quality. Particularly, the separate decoder model achieves an average improvement of $1.90$ BLEU points on English-to-Spanish ST, and $0.96$ points on English-to-French. In the case of the unified model, the gains on English-to-Spanish ST and English-to-French ST are $2.38$ and $1.84$ BLEU, respectively. In terms of improvements, the unified model is superior to the separate decoder model even with fewer parameters than the latter, indicating full parameter sharing enhances transfer learning. 
\vspace{-0.3cm}
\subsection{Asynchronous Translation}
When the asynchronous training schema is taken,  we adopt the unified model. Two settings are conducted: one is setting the $k$ to $6$ for English-to-Spanish direction and  setting $k$ to $4$ for English-to-French  while the other is setting the $k$  to $4$ for English-to-Spanish direction and setting the $k$  to $6$ for English-to-French. 

We compare asynchronous strategy with synchronous strategy and the results are shown in Table ~\ref{table3}. The asynchronous training  improves translation quality at each $k$. When setting the $k$ to $6$ for English-to-Spanish direction and  $4$ for English-to-Spanish direction, the asynchronous approach gains an improvement of $1.05$ for English-to-French translation, and $0.22$ BLEU  for English-to-Spanish translation over the synchronous training.   The other asynchronous training setting also results in  improvement of $1.78$ for English-to-Spanish translation,
and $0.42$ BLEU  for English to French translation
over the synchronous training.
These proves the effectiveness of the joint asynchronous training strategy.

It is generally acknowledged  that in the prefix-to-prefix scenario,  as we train the model to predict using source prefixes, the trained streaming model is often able  to perceive preceding context  which we call $anticipation$ \cite{DBLP:conf/acl/MaHXZLZZHLLWW19}.
We believe that with asynchronous training, the target language with  latency $k^{1}$ in the model can increase its $anticipation$ ability due to the existence of the target language with  $k^{2}$.  e.g., in joint synchronous training scenario or bilingual scenario, the target language with fixed $k$ has limited $anticipation$ ability. While in joint asynchronous training scenario, the target language with a larger $k$ plays as a transitional role, so that the target language with a smaller $k$ has an access to get the $anticipation$ ability of the target with a larger $k$. 
 
\section{Conclusions}
In this work, we have investigated joint training and decoding for multilingual end-to-end simultaneous speech translation and curated a dataset to train and evaluate the proposed methods. Experiments demonstrate that the proposed separate decoder model and unified model significantly improve translation quality of simultaneous end-to-end speech translation on a synchronous training setting. Finally, we find the joint asynchronous training method further improves translation quality.

\section{Acknowledgments}
The present research was supported by the Key Research and Development Program of Yunnan Province (Grant No. 202203AA080004). We would like to thank the anonymous reviewers for their insightful comments.

\vfill\pagebreak
\bibliographystyle{IEEEbib}
\bibliography{simple,string}

\begin{thebibliography}{10}

\bibitem{DBLP:conf/acl/SperberP20}
Matthias Sperber and Matthias Paulik,
\newblock ``Speech translation and the end-to-end promise: Taking stock of where we are,''
\newblock in {\em Proc. of ACL}, 2020.

\bibitem{DBLP:conf/acl/HuangWX21}
Wuwei Huang, Dexin Wang, and Deyi Xiong,
\newblock ``Adast: Dynamically adapting encoder states in the decoder for end-to-end speech-to-text translation,''
\newblock in {\em Proc. of ACL Findings}, 2021.

\bibitem{DBLP:conf/naacl/FiratCB16}
Orhan Firat, Kyunghyun Cho, and Yoshua Bengio,
\newblock ``Multi-way, multilingual neural machine translation with a shared attention mechanism,''
\newblock in {\em Proc. of NAACL}, 2016.

\bibitem{DBLP:conf/icassp/VuIPMSB14}
Ngoc~Thang Vu, David Imseng, Daniel Povey, Petr Motl{\'{\i}}cek, Tanja Schultz, and Herv{\'{e}} Bourlard,
\newblock ``Multilingual deep neural network based acoustic modeling for rapid language adaptation,''
\newblock in {\em Proc. of ICASSP}, 2014.

\bibitem{DBLP:conf/asru/InagumaDKW19}
Hirofumi Inaguma, Kevin Duh, Tatsuya Kawahara, and Shinji Watanabe,
\newblock ``Multilingual end-to-end speech translation,''
\newblock in {\em IEEE Automatic Speech Recognition and Understanding Workshop, ASRU 2019, Singapore, December 14-18, 2019}, 2019.

\bibitem{DBLP:conf/asru/GangiNT19}
Mattia Antonino~Di Gangi, Matteo Negri, and Marco Turchi,
\newblock ``One-to-many multilingual end-to-end speech translation,''
\newblock in {\em IEEE Automatic Speech Recognition and Understanding Workshop, ASRU 2019, Singapore, December 14-18, 2019}, 2019.

\bibitem{DBLP:conf/ijcnlp/MaPK20}
Xutai Ma, Juan~Miguel Pino, and Philipp Koehn,
\newblock ``Simulmt to simulst: Adapting simultaneous text translation to end-to-end simultaneous speech translation,''
\newblock in {\em Proc. of AACL}, 2020.

\bibitem{DBLP:conf/icassp/MaWDKP21}
Xutai Ma, Yongqiang Wang, Mohammad~Javad Dousti, Philipp Koehn, and Juan~Miguel Pino,
\newblock ``Streaming simultaneous speech translation with augmented memory transformer,''
\newblock in {\em Proc. of ICASSP}, 2021.

\bibitem{DBLP:journals/corr/BerardPSB16}
Alexandre Berard, Olivier Pietquin, Christophe Servan, and Laurent Besacier,
\newblock ``Listen and translate: {A} proof of concept for end-to-end speech-to-text translation,''
\newblock {\em CoRR}, vol. abs/1612.01744, 2016.

\bibitem{DBLP:journals/corr/abs-2009-09704}
Qianqian Dong, Mingxuan Wang, Hao Zhou, Shuang Xu, Bo~Xu, and Lei Li,
\newblock ``Listen, understand and translate": Triple supervision decouples end-to-end speech-to-text translation,''
\newblock {\em CoRR}, vol. abs/2009.09704, 2020.

\bibitem{DBLP:conf/interspeech/WeissCJWC17}
Ron~J. Weiss, Jan Chorowski, Navdeep Jaitly, Yonghui Wu, and Zhifeng Chen,
\newblock ``Sequence-to-sequence models can directly translate foreign speech,''
\newblock in {\em Proc. of INTERSPEECH}, 2017.

\bibitem{DBLP:conf/icassp/BerardBKP18}
Alexandre Berard, Laurent Besacier, Ali~Can Kocabiyikoglu, and Olivier Pietquin,
\newblock ``End-to-end automatic speech translation of audiobooks,''
\newblock in {\em Proc. of ICASSP}, 2018.

\bibitem{DBLP:conf/icassp/JiaJMWCCALW19}
Ye~Jia, Melvin Johnson, Wolfgang Macherey, Ron~J. Weiss, Yuan Cao, Chung{-}Cheng Chiu, Naveen Ari, Stella Laurenzo, and Yonghui Wu,
\newblock ``Leveraging weakly supervised data to improve end-to-end speech-to-text translation,''
\newblock in {\em Proc. of ICASSP}, 2019.

\bibitem{DBLP:conf/acl/SaleskySB19}
Elizabeth Salesky, Matthias Sperber, and Alan~W. Black,
\newblock ``Exploring phoneme-level speech representations for end-to-end speech translation,''
\newblock in {\em Proc. of ACL}, 2019.

\bibitem{DBLP:conf/emnlp/ZhangTHS20}
Biao Zhang, Ivan Titov, Barry Haddow, and Rico Sennrich,
\newblock ``Adaptive feature selection for end-to-end speech translation,''
\newblock in {\em Proc. of EMNLP Findings}, 2020.

\bibitem{DBLP:conf/interspeech/LiuXZHWWZ19}
Yuchen Liu, Hao Xiong, Jiajun Zhang, Zhongjun He, Hua Wu, Haifeng Wang, and Chengqing Zong,
\newblock ``End-to-end speech translation with knowledge distillation,''
\newblock in {\em Proc. of INTERSPEECH}, 2019.

\bibitem{DBLP:conf/acl/WangWLZY20}
Chengyi Wang, Yu~Wu, Shujie Liu, Ming Zhou, and Zhenglu Yang,
\newblock ``Curriculum pre-training for end-to-end speech translation,''
\newblock in {\em Proc. of ACL}, 2020.

\bibitem{DBLP:conf/interspeech/XueW0PG22}
Jian Xue, Peidong Wang, Jinyu Li, Matt Post, and Yashesh Gaur,
\newblock ``Large-scale streaming end-to-end speech translation with neural transducers,''
\newblock in {\em Proc. of INTERSPEECH}, 2022.

\bibitem{DBLP:conf/coling/LePWGSB20}
Hang Le, Juan~Miguel Pino, Changhan Wang, Jiatao Gu, Didier Schwab, and Laurent Besacier,
\newblock ``Dual-decoder transformer for joint automatic speech recognition and multilingual speech translation,''
\newblock in {\em Proc. of COLING}, 2020.

\bibitem{di-gangi-etal-2019-must}
Mattia~A. Di~Gangi, Roldano Cattoni, Luisa Bentivogli, Matteo Negri, and Marco Turchi,
\newblock ``{M}u{ST}-{C}: a {M}ultilingual {S}peech {T}ranslation {C}orpus,''
\newblock in {\em Proc. of NAACL}, 2019.

\bibitem{long-etal-2020-ted}
Wanqiu Long, Bonnie Webber, and Deyi Xiong,
\newblock ``{TED}-{CDB}: A large-scale {C}hinese discourse relation dataset on {TED} talks,''
\newblock in {\em Proceedings of the 2020 Conference on Empirical Methods in Natural Language Processing (EMNLP)}, Bonnie Webber, Trevor Cohn, Yulan He, and Yang Liu, Eds., Online, Nov. 2020, pp. 2793--2803, Association for Computational Linguistics.

\bibitem{long2020shallowdiscourseannotationchinese}
Wanqiu Long, Xinyi Cai, James E.~M. Reid, Bonnie Webber, and Deyi Xiong,
\newblock ``Shallow discourse annotation for chinese ted talks,'' 2020.

\bibitem{DBLP:conf/acl/PapineniRWZ02}
Kishore Papineni, Salim Roukos, Todd Ward, and Wei{-}Jing Zhu,
\newblock ``Bleu: a method for automatic evaluation of machine translation,''
\newblock in {\em Proc. of ACL}, 2002.

\bibitem{DBLP:conf/acl/MaHXZLZZHLLWW19}
Mingbo Ma, Liang Huang, Hao Xiong, Renjie Zheng, Kaibo Liu, Baigong Zheng, Chuanqiang Zhang, Zhongjun He, Hairong Liu, Xing Li, Hua Wu, and Haifeng Wang,
\newblock ``{STACL:} simultaneous translation with implicit anticipation and controllable latency using prefix-to-prefix framework,''
\newblock in {\em Proc. of ACL}, 2019.

\bibitem{DBLP:journals/corr/ChoE16}
Kyunghyun Cho and Masha Esipova,
\newblock ``Can neural machine translation do simultaneous translation?,''
\newblock {\em CoRR}, 2016.

\end{thebibliography}

\clearpage
\end{document}